\documentclass[letterpaper, 10 pt, conference]{ieeeconf}  

\usepackage{amsmath}
\usepackage{graphicx}
\usepackage{float}
\usepackage{booktabs}
\usepackage{siunitx}
\usepackage{caption}
\usepackage{ragged2e}
\usepackage{multirow}
\usepackage{cancel}
\usepackage{subcaption}
\usepackage{soul}
\usepackage{amsmath, amssymb}
\usepackage{algorithm}
\usepackage{algpseudocode}
\usepackage{pifont} 
\usepackage{hyperref}
\usepackage{multicol}
\usepackage{xcolor}
\usepackage{fancyhdr}

\definecolor{green_ad}{HTML}{46b361}
\definecolor{red_ad}{HTML}{E74C3C}
\definecolor{blue_ad}{HTML}{2E86C1}
\definecolor{lightgrey}{gray}{0.4} 
\newcommand{\ci}[1]{\textcolor{lightgrey}{_{\pm#1}}}

\newcommand{\cmark}{\ding{51}}  
\newcommand{\xmark}{\ding{55}}  

\usepackage[font=small,labelfont=bf, hypcap=false]{caption}

\title{\LARGE \bf Hebbian Attractor Networks for Robot Locomotion}
\author{
Alexander Dittrich, Fuda van Diggelen and Dario Floreano,~\textit{Fellow,~IEEE}\\
\small Laboratory of Intelligent Systems, EPFL, Switzerland\\
\small \texttt{\{alexander.dittrich,fuda.vandiggelen,dario.floreano\}@epfl.ch}\\%
}

\usepackage{fancyhdr}
\begin{document}

\twocolumn[{%
\renewcommand\twocolumn[1][]{#1}%
\maketitle
\begin{center}
    \centering
    \captionsetup{type=figure}
    \includegraphics[width=\linewidth]{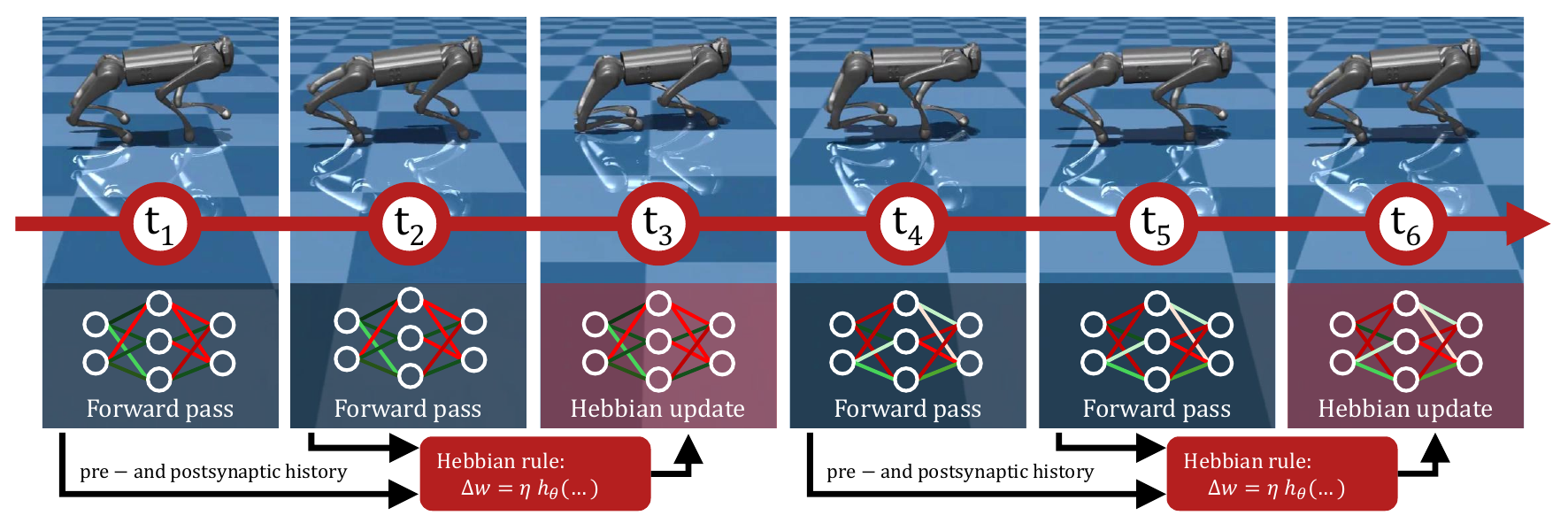}
    \captionof{figure}{Hebbian neural networks are updated during the lifetime based on correlations between pre- and postsynaptic activations and do not rely on gradient information. By introducing a slower rate of Hebbian updates than the action inference or averaging of the synaptic history, we can induce different attractor dynamics in weight space.}
    \label{fig:hebbian_learning}
\end{center}
}]
\raggedbottom
\captionsetup{font=small}



\begin{abstract}
Biological neural networks continuously adapt and modify themselves in response to experiences throughout their lifetime—a capability largely absent in artificial neural networks. Hebbian plasticity offers a promising path toward rapid adaptation in changing environments. Here, we introduce Hebbian Attractor Networks (HAN), a class of plastic neural networks in which local weight update normalization induces emergent attractor dynamics. Unlike prior approaches, HANs employ dual-timescale plasticity and temporal averaging of pre- and postsynaptic activations to induce either co-dynamic limit cycles or fixed-point weight attractors. Using simulated locomotion benchmarks, we gain insight into how Hebbian update frequency and activation averaging influence weight dynamics and control performance. Our results show that slower updates, combined with averaged pre- and postsynaptic activations, promote convergence to stable weight configurations, while faster updates yield oscillatory co-dynamic systems. We further demonstrate that these findings generalize to high-dimensional quadrupedal locomotion with a simulated Unitree Go1 robot. These results highlight how the timing of plasticity shapes neural dynamics in embodied systems, providing a principled characterization of the attractor regimes that emerge in self-modifying networks.
\end{abstract}

\renewcommand\thefootnote{}
\footnotetext{
Code and supplementary material at \url{https://github.com/alexanderdittrich/hebbian-attractor-networks}}
\addtocounter{footnote}{-1}

\renewcommand\thefootnote{} 
\footnotetext{
All authors are affiliated with the Laboratory of Intelligent Systems, École Polytechnique Fédérale de Lausanne (EPFL), Lausanne, Switzerland.
}
\addtocounter{footnote}{-1}

\thispagestyle{fancy}


\section{Introduction}

Robot learning approaches follow a train-deploy paradigm: a neural network policy is trained in simulation to master a pre-defined task, by optimizing a static set of weight values. During deployment, these weights remain fixed, preventing adaptation to novel or changing conditions~\cite{lee2020learning, li2024reinforcement, ha2024learning}. However, real-world environments are dynamic—sensors drift, terrain conditions vary, and actuator responses degrade. This brittleness contrasts sharply with biological nervous systems, which leverage continuous synaptic plasticity to self-modify throughout the organism's lifetime. Inspired by the adaptivity of biological systems, we investigate continuously self-modifying networks in the form of Hebbian learning.

Hebbian neural networks (HNNs) display continuous adaptation of connection weights according to the co-activation of pre- and postsynaptic neurons \cite{hebb1949organization}. When integrated into artificial neural networks, Hebbian plasticity introduces dynamic, self-modifying behaviors during deployment. Prior work has shown that Hebbian learning can produce effective controllers~\cite{najarro2020meta, yaman2021evolving, ferigo2022evolving, schmidgall2020, leung2025bio, van2025emergent}.
However, how weight changes interact with control, stabilize behavior, and prevent runaway dynamics remain poorly understood. 

In this work, we introduce Hebbian Attractor Networks (HANs)—a class of Hebbian neural networks where weight normalization induces emergent attractor dynamics. We show that max-normalized plasticity does more than prevent unbounded weight growth: it enables structured, recurring weight behaviors such as limit cycles and fixed-point attractors. These attractor dynamics are shaped by the temporal structure of the plasticity—specifically, the frequency of Hebbian updates and the averaging window of synaptic activations (\autoref{fig:hebbian_learning}). Our method draws inspiration from the multi-timescale architecture of biological systems \cite{bittner2017behavioral, gerstner2018eligibility}. We implement Hebbian updates that are decoupled from the control loop and based on temporally averaged activation traces. These modifications result in qualitatively distinct weight dynamics. Fast updates promote oscillatory co-dynamic systems, where weight changes and motor control are tightly coupled. Slower updates with longer averaging windows induce stable fixed-points in weight space. Using standard locomotion benchmarks, we show that HANs support qualitatively distinct 
attractor regimes---fixed points and limit cycles---whereas prior Hebbian approaches 
with synchronized updates produce only co-dynamic limit 
cycles~\cite{najarro2020meta, leung2025bio}. By leveraging dual-timescale plasticity and activation averaging, HANs enable both co-dynamic and fixed-point attractor regimes—offering improved robustness and adaptability in robot locomotion. Here, plasticity is considered both in its role in meta-learning and as a mechanism that can give rise to characteristic network dynamics. 

The contribution of this paper is three-fold:
(i) We provide a systematic analysis of HANs on continuous-control locomotion tasks, situating them relative to static evolutionary controllers and gradient-based RL baselines.
(ii) We characterize the emergence of distinct attractor regimes in weight space, showing how update frequency and activation averaging shape the onset of oscillatory versus fixed-point dynamics.
(iii) We demonstrate that the resulting attractor dynamics support adaptive control beyond toy benchmarks, scaling to quadrupedal locomotion on a simulated Unitree Go1 robot.


\section{Related Work}
\label{sec:citations}

\textbf{Self-modifying networks and meta-learning:} 
From a wider perspective, Hebbian learning can be considered as a meta-learning approach. Other meta-learning methods like MAML \cite{finn2017model} or fast weights \cite{schmidhuber1992learning, ba2016using} enable rapid adaptation by performing gradient-based updates during deployment. These systems require explicit error signals and backpropagation to fine-tune parameters in real time. In contrast, Hebbian learning provides a bio-inspired alternative: adaptation occurs through local, activity-dependent plasticity without relying on global error signals or gradient computations.

\textbf{Hebbian learning and weight dynamics:} Hebbian plasticity utilizes local synaptic updates dependent solely on the activities of pre- and postsynaptic neurons, making it computationally simple and biologically plausible. Extensive research on Hebbian weight dynamics, such as Oja’s rule~\cite{oja1982simplified} and the BCM theory~\cite{bienenstock1982theory}, has provided insights into how synaptic weights can self-modify toward stable configurations. However, these classical models primarily address static tasks or stationary input distributions, whereas in our work, we focus on dynamic, continuously evolving environments. While in its principal form (a product of pre- and postsynaptic activations), Hebbian learning can be optimized using backpropagation~\cite{miconi2018differentiable, miconi2018backpropamine}, other work has explored the evolution of more expressive Hebbian rule formulations~\cite{floreano2000evolutionary, soltoggio2008evolutionary, pedersen2021evolving, yaman2021evolving}. Evolutionary algorithms are capable of exploring high-dimensional search spaces without imposing restrictive constraints on rule formulation~\cite{salimans2017evolution, miikkulainen2025neuroevolution}. Several studies~\cite{schmidgall2020, najarro2020meta, leung2025bio, van2025emergent} commonly use a generalized Hebbian formulation, known as the ABCD rule, where synaptic weight updates depend linearly on combinations of pre- and postsynaptic neuron activations. To mitigate unstable weight growth, Max Normalization (MN) by rescaling the weights has been employed, improving performance and stability~\cite{leung2025bio}. Najarro et al. \cite{najarro2020meta} demonstrated that evolved Hebbian rules can enable networks to rapidly recover from physical damage, effectively self-organizing during deployment. Ferigo et al. \cite{ferigo2022evolving} further showed in voxel-based robots the emergence of oscillatory weights with MN enabled, indicating a co-dynamic system in which the plastic weights act more as intrinsic pattern generators rather than mechanisms of learning. While prior work has demonstrated the utility of Hebbian learning in adaptive control, the role of update timing and temporal structure in shaping  converging versus oscillatory behaviors in embodied systems remains largely unexplored — a gap we address through the framework of HANs.



\section{Hebbian Attractor Networks}
\label{sec:hebbian_learning}



\subsection{Hebbian Neural Networks}

The NNs used in this paper have a feedforward architecture consisting of $L$ layers. Let $\boldsymbol{x}^{(k)}(t)$ denote the vector of activations at layer $k$ and time $t\in\mathbb{N}$, and let $W^{(k)}(t)$ be the matrix of weights connecting layer $k-1$ to $k$. The network receives a state $\boldsymbol{s}_t$ and produces a deterministic action $\boldsymbol{a}_t$ as follows:
\begin{align}
\boldsymbol{x}^{(0)}(t) &= \boldsymbol{s}_t, \\
\boldsymbol{x}^{(k)}(t) &= \varphi\left(W^{(k)}(t) \boldsymbol{x}^{(k-1)}(t)\right), \quad \text{for } k = 1, \dots, L, \\
\boldsymbol{a}_t &= \boldsymbol{x}^{(L)}(t),
\end{align}
where $\varphi$ is a nonlinear activation function (in our case, $\tanh$). The synaptic weights $W^{(k)}(t)$ are updated online during deployment via local Hebbian rules. Each connection $(i,j)$ in layer $k$ is associated with a parameterized Hebbian update function $h_{\theta_{ij}^{(k)}}$ and a local learning rate $\eta_{ij}^{(k)}$. The scalar Hebbian update at discrete time step $t$ is computed as:
\begin{equation}
\Delta w_{ij}^{(k)}(t) = \eta_{ij}^{(k)} \cdot h_{\theta_{ij}^{(k)}}\left(\bar{x}_j^{(k-1)}(t), \bar{x}_i^{(k)}(t)\right),
\end{equation}
where the presynaptic activation $\bar{x}_j^{(k-1)}(t)$ and postsynaptic activation $\bar{x}_i^{(k)}(t)$ are computed as a moving average (MA) of length $M$ of the neuron's activation values:
\begin{align}\label{eq:MA}
\bar{x}_i^{(k)}(t)   &= \frac{1}{M} \sum_{m=0}^{M-1} x_i^{(k)}(t - m) 
\end{align}

The update function $h_{\theta_{ij}^{(k)}}$ follows the commonly used generalized ABCD rule and is defined as:
\begin{multline}
\label{eq:ghl}
h_{\theta_{ij}^{(k)}}(x_j^{(k-1)}, x_i^{(k)}) = \\
a_{ij}^{(k)} x_j^{(k-1)} x_i^{(k)} 
+ b_{ij}^{(k)} x_j^{(k-1)}
+ c_{ij}^{(k)} x_i^{(k)} 
+ d_{ij}^{(k)},
\end{multline}
where $\theta_{ij}^{(k)} = (a_{ij}^{(k)}, b_{ij}^{(k)}, c_{ij}^{(k)}, d_{ij}^{(k)})$ are the Hebbian coefficients of the update function. Both these coefficients and the associated local learning rates $\{ \boldsymbol{\theta}, \boldsymbol{\eta} \}$ are optimized using Evolutionary Strategies (ES) \cite{rechenberg1978evolutionsstrategien}. 
After each Hebbian update, we apply a layerwise MN to prevent unbounded growth and enable more complex weight dynamics \cite{ferigo2022evolving, leung2025bio}. HNNs that utilize MN are denoted as HANs. Namely, for each layer $k$, we scale weights by their maximum absolute value:
\begin{equation}
\label{eq:ws}
w_{ij}^{(k)}(t+1) \leftarrow \frac{w_{ij}^{(k)}(t+1)}{\max_{i,j} |w_{ij}^{(k)}(t+1)|}
\end{equation}

\subsection{Dual-timescale in Hebbian Attractor Networks}
\begin{figure}[H] 
    \centering
    \includegraphics[width=0.98\linewidth]{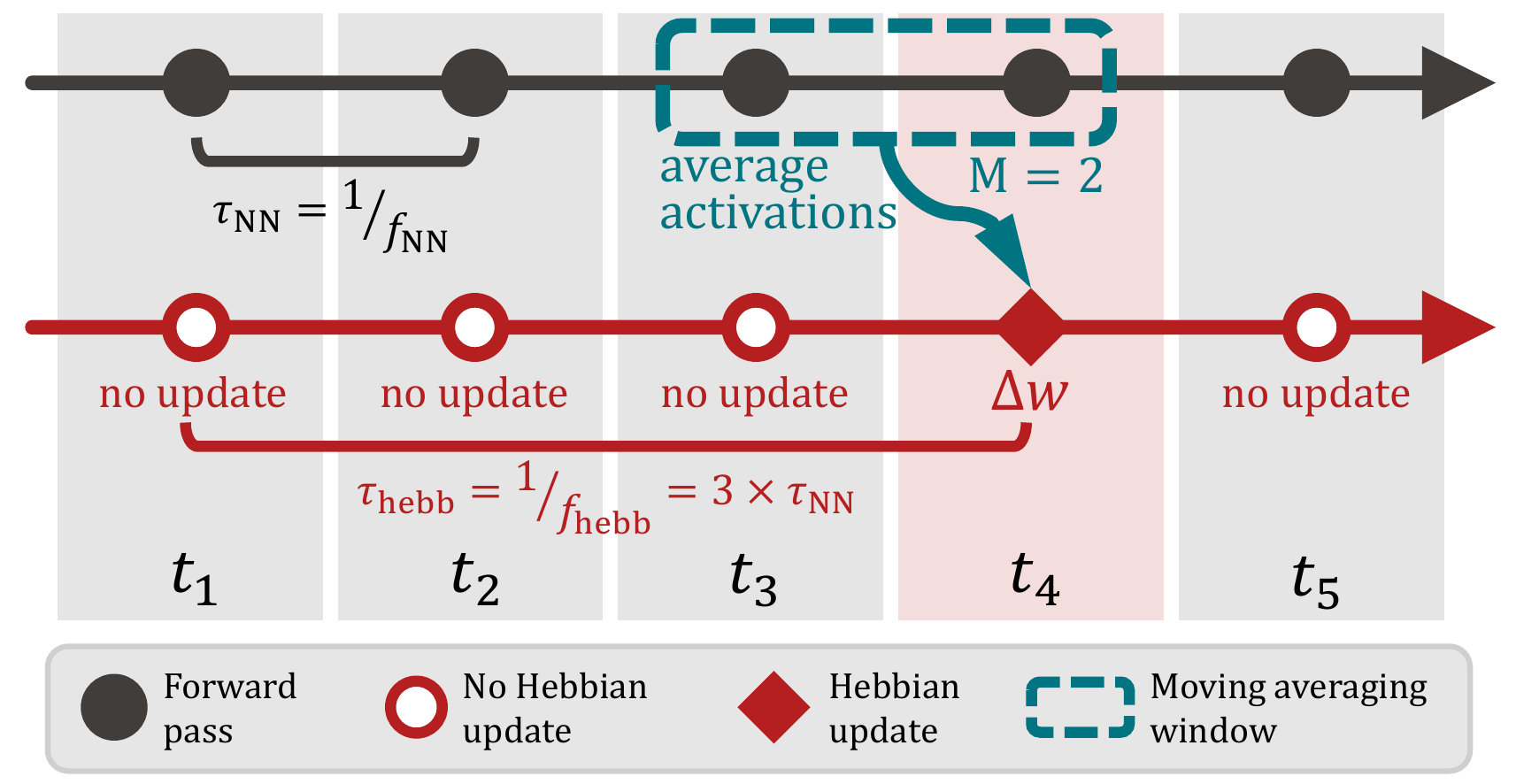}
    \caption{Dual-timescale structure of HNN. Hebbian updates are executed at a lower frequency than the forward pass (here, $\tau_{\text{hebb}} = 3\times \tau_{\text{NN}}$), and the moving average of pre- and postsynaptic activations is applied (here, $M=2$).}
    \label{fig:illustration_update_frequency}
\end{figure}

Unlike previous works \cite{najarro2020meta, leung2025bio}, we do not apply Hebbian updates at every time step. Instead, we define a fast NN controller frequency $f_{\text{NN}}$ and a slow Hebbian update frequency $f_{\text{hebb}}$ and apply updates every $\tau_{\text{hebb}} = \left\lfloor f_{\text{NN}} / f_{\text{hebb}} \right\rfloor$ steps: 
\begin{equation}
w_{ij}^{(k)}(t+1) =
\begin{cases}
w_{ij}^{(k)}(t) + \Delta w_{ij}^{(k)}(t), & \text{if } t \bmod \tau_{\text{hebb}} = 0, \\
w_{ij}^{(k)}(t), & \text{otherwise}.
\end{cases}
\end{equation}
This decoupling introduces a dual-timescale structure, as displayed in \autoref{fig:illustration_update_frequency}. We aggregate pre- and postsynaptic activations over multiple forward passes and compute a moving average (\autoref{eq:MA}). This averaging enables the learning rule to operate on smoothed activation signals that reflect longer temporal dynamics rather than relying solely on instantaneous activations.

In the following, we denote the variations of HANs in $(M=...)$ superscript for the moving average and the timescale in $(f_{\text{NN}}= ... \times f_{\text{hebb}})$ subscript, e.g. HAN$^{M=10}_{f_{\text{NN}}=4\times f_{\text{hebb}}}$.

\section{Experimental Results}
\label{sec:result}
We now investigate empirically how different structural components of Hebbian learning affect performance and weight dynamics.

\begin{table}[th]
\vspace{2mm}
\centering
\caption{Overview of tested HAN conditions.}
\begin{tabular}{@{}lccccc@{}}
\toprule
Condition & MN & $M$ & $f_{\text{NN}} / f_{\text{hebb}}$ \\
\midrule
(A) HNN & \xmark & 1 & 1 \\ \addlinespace[2pt]
(B) HAN$^{M=1}_{f_{\text{NN}}=f_{\text{hebb}}}$ & \cmark & 1 & 1 \\ \addlinespace[2pt]
(C) HAN$^{M=1}_{f_{\text{NN}}=4\times f_{\text{hebb}}}$ & \cmark & 1 & 4 \\ \addlinespace[2pt]
(D) HAN$^{M=10}_{f_{\text{NN}}=f_{\text{hebb}}}$ & \cmark & 10 & 1 \\ \addlinespace[2pt]
(E) HAN$^{M=10}_{f_{\text{NN}}=4\times f_{\text{hebb}}}$ & \cmark & 10 & 4 \\
\bottomrule
\end{tabular}
\label{tab:ablation_conditions}
\vspace{-3mm}
\end{table}

 Specifically, we perform an ablation on the effects of MN, dual-timescale plasticity ($f_{\text{NN}}>f_{\text{hebb}}$), and MA ($M>1$) across a selected set of Hebbian learning conditions.  \autoref{tab:ablation_conditions} lists the ablation conditions of HANs.
For instance, the work of Najarro et al. \cite{najarro2020meta} uses the HAN  condition~(B) (HAN$^{M=1}_{f{\text{NN}}=f_{\text{hebb}}}$) with MN, without MA ($M=1$), and a single-timescale ($f_{\text{NN}} = f_{\text{hebb}}$).

\subsection{Hebbian learning conditions influence performance}
\label{sec:benchmarking}

We evaluate HNNs and HANs on the standard continuous-control benchmarks from Gymnasium~\cite{towers2024gymnasium}, namely Swimmer-v5, HalfCheetah-v5, Hopper-v5, and Walker2d-v5. These environments provide widely used reference tasks for assessing the utility of learning algorithms in locomotion and allow us to situate the behavior of Hebbian plasticity in relation to established methods. For all Hebbian architectures, we employ a compact network with a single hidden layer of $16$ units, resulting in $800$–$1840$ parameters depending on the task. Initial weights $\boldsymbol{W}$ are sampled from $\mathcal{U}(-0.1,0.1)$, while Hebbian coefficients and learning rates ${\boldsymbol{\theta},\boldsymbol{\eta}}$ are drawn from $\mathcal{U}(-1.0,1.0)$. Optimization is performed with ES using evosax~\cite{lange2023evosax}. The most important hyperparameters include a population size of $128$, mutation rate of $0.5$, and selection ratio of \SI{10}{\percent} over $1000$ generations ($500$ for Swimmer). Each setting is evaluated four times per run, and results are averaged over $16$ independent seeds (random initializations of network weights and environment seeds).

\begin{figure*}[htb]
    \centering
    \includegraphics[width=0.92\linewidth]{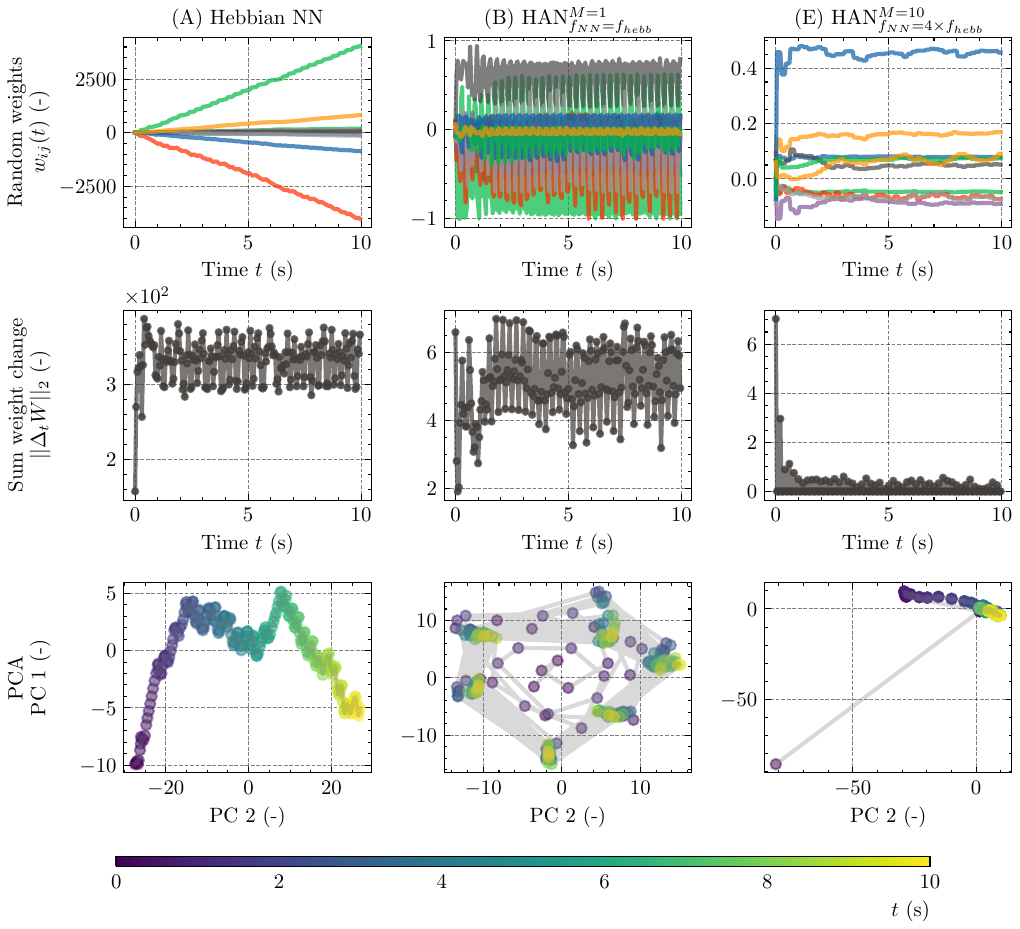}
    \caption{Analysis of weight dynamics under three Hebbian learning conditions (A, B, E). The first row shows trajectories of 10 randomly selected plastic weights over time. The second row presents the total network plasticity, quantified as the summed $\ell_2$-norm of weight changes at each timestep. The third row depicts the evolution of plastic weights projected onto the first two principal components via Principal component analysis (PCA).}
    \label{fig:cheetah_weight_dynamics}
\end{figure*}

\begin{table}[h!]
\footnotesize
\centering
\setlength{\tabcolsep}{3pt}
\renewcommand{\arraystretch}{1.2} 
\caption{Average accumulated episodic return / fitness on Gymnasium locomotion benchmarks ($n=16$ for \textcolor{green_ad}{HNN/HAN} and \textcolor{blue_ad}{MLP/GRU}; $n=10$ for \textcolor{red_ad}{PPO}), where $\pm$ captures a \SI{95}{\percent} confidence interval. \textcolor{blue_ad}{MLP$^{\dagger}$} omits bias terms to match the parameter count of HNNs/HANs. \textcolor{red_ad}{PPO$^\ddagger$} matches the number of optimization parameters of HNNs/HANs.}
\label{tab:benchmark_results}
\begin{tabular}{@{}lcccc@{}}
\toprule
\multirow{2}{*}{\textbf{Algorithm}} & \multicolumn{4}{c}{\textbf{Fitness (-)}} \\
 & Swimmer & Hopper & HalfCheetah & Walker2D \\ \midrule
\textcolor{red_ad}{PPO$^\ddagger$} & $134.4 \ci{1.8}$ & $2641.2 \ci{1133.1}$ & $3722.9 \ci{1842.4}$ & $4022.8 \ci{2253.7}$ \\
\textcolor{red_ad}{PPO}        & $135.7 \ci{3.5}$ & $4068.8 \ci{179.5}$  & $9644.7 \ci{1084.4}$ & $7947.8 \ci{899.6}$ \\ \midrule
\textcolor{blue_ad}{MLP$^\dagger$} & $361.4 \ci{0.4}$ & $3470.9 \ci{195.2}$ & $3496.7 \ci{382.6}$ & $3292.3 \ci{121.6}$ \\
\textcolor{blue_ad}{MLP}        & $361.9 \ci{0.3}$ & $4186.0 \ci{129.8}$ & $3168.7 \ci{115.5}$ & $4310.9 \ci{339.1}$ \\
\textcolor{blue_ad}{GRU}        & $363.7 \ci{0.4}$ & $3676.8 \ci{159.5}$ & $4454.6 \ci{413.5}$ & $4066.1 \ci{418.9}$ \\ \midrule
\textcolor{green_ad}{(A) HNN}  & $359.5 \ci{0.6}$ & $2466.0 \ci{382.1}$ & $3370.3 \ci{601.7}$ & $3033.3 \ci{127.4}$ \\
\textcolor{green_ad}{(B) HAN}  & $363.6 \ci{0.2}$ & $3713.1 \ci{322.6}$ & $6385.3 \ci{601.9}$ & $4532.5 \ci{831.1}$ \\
\textcolor{green_ad}{(C) HAN}  & $363.7 \ci{0.2}$ & $3697.0 \ci{276.3}$ & $5055.5 \ci{762.6}$ & $4261.1 \ci{417.1}$ \\
\textcolor{green_ad}{(D) HAN}  & $363.9 \ci{0.2}$ & $3729.7 \ci{330.2}$ & $7239.6 \ci{430.2}$ & $4548.5 \ci{492.4}$ \\
\textcolor{green_ad}{(E) HAN}  & $363.9 \ci{0.3}$ & $4100.0 \ci{172.3}$ & $7394.0 \ci{301.5}$ & $4046.9 \ci{280.9}$ \\ \bottomrule
\end{tabular}
\end{table}

To isolate the contribution of plasticity, we compare HANs against several directly evolved static baselines using the same ES. These include (i) MLP controllers without plasticity but matched in parameter count (without bias terms), (ii) MLP and GRU controllers with bias parameters, where the GRU provides behavioral adaptation through recurrent hidden state rather than synaptic plasticity. This design allows us to assess how much of the observed performance stems from within-lifetime Hebbian adaptation, beyond what can be captured by recurrence or direct optimization.

For additional context, we also report results with PPO \cite{schulman2017proximal}, using Stable-Baselines3~\cite{raffin2021stable}. Two architectures are considered: (a) a “tiny” variant matching the HANs (single hidden layer with $16$ units for the actor), and (b) the default PPO architecture with two $64$-unit hidden layers for the actor. In both cases, the critic uses the default two-layer $64$-unit architecture. PPO is trained for $80$ million environment steps with default hyperparameters, and results are averaged over $10$ seeds. These PPO runs provide reference points, not direct benchmarks, as the optimization algorithms and inductive biases differ substantially. 

\begin{figure*}
    \centering
    \includegraphics[width=0.94\linewidth]{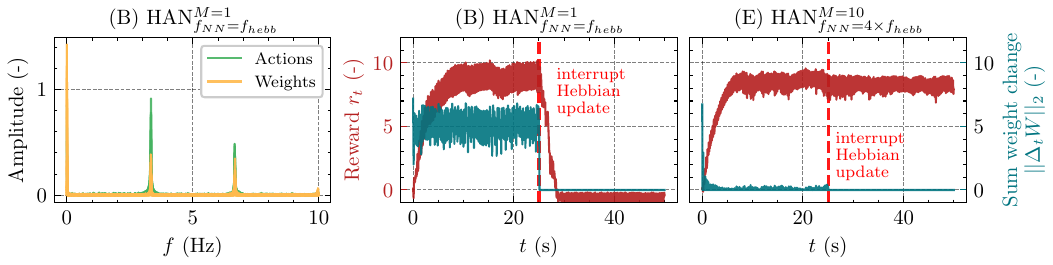}
    \caption{The non-zero frequencies of action and weight signals exhibit peaks at approximately \SI{4}{\hertz} and \SI{8}{\hertz}. Interrupting the Hebbian update causes a pronounced drop in step rewards during deployment for the HAN condition (B). The HNN in condition (E) shows a minor effect in step rewards.}
    \label{fig:weight_freeze}
\end{figure*}

\autoref{tab:benchmark_results} reports mean performance across conditions. The ablation study shows that adding max-normalization (MN) substantially improves performance across all tasks. Incorporating either activation averaging, dual timescales, or both yields further gains. As a result, HANs outperform directly comparable static MLPs without bias, and in most cases also surpass MLPs and GRUs with bias, except for Hopper where static baselines remain competitive. 
When compared to PPO, HANs achieve comparable returns on simpler environments such as Swimmer, while gradient-based PPO with its larger default architecture attains the highest scores on Hopper, HalfCheetah, and Walker2D. These differences reflect the respective strengths of the methods: PPO leverages gradient updates and larger networks, whereas HANs exploit within-lifetime plasticity to enrich the expressivity of compact architectures. Importantly, HANs demonstrate that Hebbian rules can be harnessed to improve performance beyond what is achievable with static evolution or direct gradient optimization on the same small architectures.

\subsection{Fixed-point attractors emerge with decoupled Hebbian updates and averaging of activations}
\label{sec:oscillations}

For further investigation of the attractor dynamics, we select the best-performing Hebbian coefficients in the HAN conditions (A), (B) and (E) of the HalfCheetah task. \autoref{fig:cheetah_weight_dynamics} shows the weight dynamics during a single lifetime over \SI{10}{\second}. The first row depicts $10$ randomly selected plastic weights. For the second row, we compute the elementwise $\ell_2$ distances of the weight matrices between the current $t_n$ and the previous time step $t_{n-1}$, and we sum them over the network. This metric gives an indication of overall network plasticity, where a value of 0 signifies no change in network weights. In the third row, we use a PCA to reduce the plastic weights at each time step to two dimensions. 

For reference, HNN in condition (A) shows the expected unbounded growth: individual weights diverge and no distinct attractor behavior emerges, as confirmed by the PCA embedding. Similarly, we observe sustained high plasticity in HANs under condition (B), with single weights oscillating periodically. The plastic weights remain within a $(-1, 1)$-bound. In the PCA embedding, we observe that the plastic weights form an oscillatory pattern, where the principal components jump between several states. In condition (E), the total plastic weight change decreases after a short phase, remaining close to $0$ thereafter. We observe that single weights remain constant after an initial jump. In the PCA embedding, the compressed weights remain within a region of the weight space after a strong initial change.

\subsection{Attractor analysis}
\label{sec:indepth}

The results of \autoref{sec:oscillations} show the onset of distinct weight dynamics conditioned on the different HAN conditions. We investigate the design of these different weight dynamics through additional ablation studies by altering Hebbian update frequency $f_{\text{hebb}}$ and averaging window length $M$. We use HANs in condition (E) for the HalfCheetah task and perform $16$ randomly initialized evolutions for each combination of window lengths $M=1, 2, 4, 10, 20$ and Hebbian update frequencies $f_{\text{hebb}}=1, 5, 10, 20$. 

\begin{figure}[ht]
    \centering
    \includegraphics[width=\linewidth]{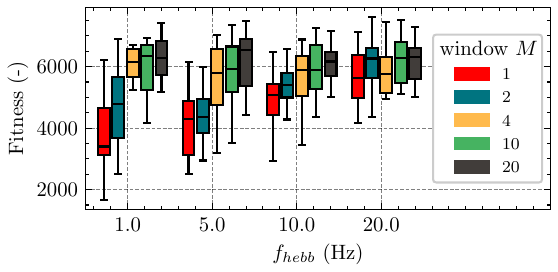}
    \captionof{figure}{Fitness at 1000 generations with different $f_{\text{hebb}}$ and $M$ ($n=16$). A window length of $M=1$ corresponds to no moving average, and $f_{\text{hebb}}=20$ corresponds to no reduction in update frequency.}
    \label{fig:smoothing_frequencies}
\end{figure}

\autoref{fig:smoothing_frequencies} shows the fitness of the best performing individual at the final generation over different random seeds.

\begin{figure*}
    \centering
    \includegraphics[width=\linewidth]{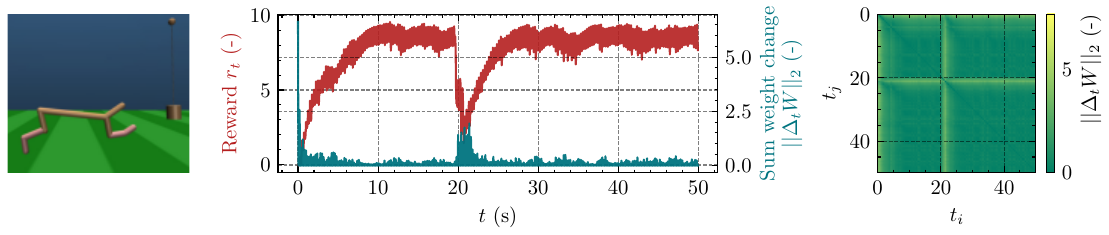}
    \caption{Stepwise rewards summed weight changes during a \SI{50}{\second} deployment. A collision around \SI{19}{\second} causes a drop in reward, followed by recovery. Weight changes decrease over time, with a brief spike after the collision. The network returns to a similar weight configuration after the disturbance, indicating a stable attractor.}
    \label{fig:perturbation}
\end{figure*}


To assess whether a HAN converges to a fixed-point attractor, we track the weight change over time in the early and late phase of an episode. We compare the difference of the total weight change $\Delta_t W = \sum_k \| W^{(k)}_t - W^{(k)}_{t-1} \|_2$, where early denotes the first \SI{5}{\percent} of the rollout, and late the remaining rollout. We classify the HAN condition as converged if $\overline{\Delta W}_{\text{late}} < \rho \cdot  \overline{\Delta W}_{\text{early}}$. For robustness, we added the threshold factor $\rho=0.9$ and evaluate $3$ population members at the last generation of $16$ seeds for $30$ independent rollouts. Each combination of $f_{\text{hebb}}$ and $M$ is evaluated $1440$ times.


\begin{table}[htb]
    \centering
    \caption{Ratio of HANs satisfying the convergence criterion for different $f_{\text{hebb}}$ and $M$ values over $1440$ different evaluations.}
    \label{tab:converging_weights}
    \begin{tabular}{lcccc}
        \toprule
        &\multicolumn{4}{c}{Hebbian update frequency $f_{\text{hebb}}$} \\
        \cmidrule(lr){2-5}
        $M$ & \SI{1}{\hertz} & \SI{5}{\hertz} & \SI{10}{\hertz} & \SI{20}{\hertz} \\
        \midrule
        1  & \SI{99.9}{\percent} & \SI{64.4}{\percent} & \SI{21.9}{\percent} & \SI{21.4}{\percent} \\
        2  & \SI{99.9}{\percent} & \SI{64.9}{\percent} & \SI{16.1}{\percent} & \SI{17.8}{\percent} \\
        4  & \SI{99.9}{\percent} & \SI{81.7}{\percent} & \SI{55.0}{\percent} & \SI{36.6}{\percent} \\
        10 & \SI{99.9}{\percent} & \SI{94.4}{\percent} & \SI{75.7}{\percent} & \SI{66.0}{\percent} \\
        20 & \SI{99.9}{\percent} & \SI{100.0}{\percent} & \SI{100.0}{\percent} & \SI{92.9}{\percent} \\
        \bottomrule
    \end{tabular}  
\end{table}

The results show that convergence to a fixed-point attractor depends on the choice of $f_{\text{hebb}}$ and $M$. To achieve converging plastic weights, either low Hebbian update frequencies $f_{\text{hebb}}$, large average window lengths $M$ or a combination of both is effective. Low $f_{\text{hebb}}$ consistently results in converging weights; however, it has negative impacts on performance compared to higher $f_{\text{hebb}}$. We found that $M\geq4$ mitigated the decrease in performance while maintaining the convergence properties.

We continue by analyzing the different properties of the weight attractors. A Fourier analysis on $30$ randomly selected plastic weights $w_{ij}^{(k)}$ of the HANs in condition (B) shows that the dynamics of the weights are in tune with the locomotion pattern of the agent. The main non-zero frequencies of the plastic weights and the action signal are highly correlated (\autoref{fig:weight_freeze}, left). When we pause Hebbian updates during the lifetime, we see that the oscillatory attractor drives a vital part of locomotion as the agent is unable to move (\autoref{fig:weight_freeze} middle, a substantial drop in reward after $t=20$). This is not the case for the condition (E) which indicates a more robust attractor behavior. 

The stability of the attractor is analyzed in a final perturbation experiment. Here, we randomly place a hanging pendulum with a mass of \SI{50}{\kilogram} in the locomotion path as a force perturbation (\autoref{fig:perturbation}, left). This condition has not been part of the training process; it thus requires online adaptation of the network. 
HANs in condition (E) show high plasticity at the start of periodic locomotion until a steady-state periodic gait is reached. Perturbations from this gait result in an increase in plasticity. \autoref{fig:perturbation} (right) shows the summed $\ell_2$ distance of weight snapshots between each timestep of the lifetime. This allows the comparison of all weight snapshots during the lifetime. The summed distance between plastic weights before and after the perturbation is close to zero, indicating the existence of a single fixed-point attractor and therefore a single weight configuration the network converges to.

\subsection{Unitree Go1 locomotion}
\label{sec:quad}
To assess whether our findings generalize to higher-dimensional systems, we extend to quadrupedal locomotion using a Unitree Go1 robot simulated in MuJoCo XLA (MJX)~\cite{todorov2012mujoco}, modified from the Joystick task in MuJoCo Playground~\cite{zakka2025mujoco}. The reward consists of a Gaussian tracking term for target forward velocities of \SIlist{1.0;1.5;2.0}{\meter\per\second} and a Gaussian term for maintaining zero yaw. We employ OpenAI-ES~\cite{salimans2017evolution} with a population size of $512$, exponential decay schedules for learning rate ($0.1$, decay $0.999$) and mutation rate ($0.2$, decay $0.995$) over $500$ generations, with four repeats per evaluation and $1000$-step episodes at $f_{\text{NN}}=\SI{50}{\hertz}$.

\begin{figure}[bht]
    \centering
    \includegraphics[width=0.96\linewidth]{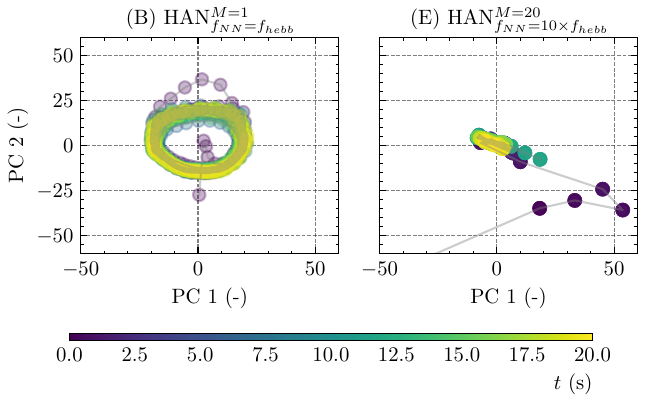}
    \caption{PCA of HANs in conditions (B) and (E).}
    \label{fig:go1_pca}
\end{figure}

The observation space includes trunk angular and linear velocities, joint angles and velocities, and the gravity vector. The network outputs desired joint angles for each leg's hip, thigh, and calf motors, tracked by a PD controller ($k_p=50$, $k_d=1$) at \SI{1000}{\hertz}. We train $10$ randomly initialized HANs per target velocity under condition (B) (HAN$^{M=1}_{f_{\text{NN}}=f_{\text{hebb}}}$) and condition (E) (HAN$^{M=20}_{f_{\text{NN}}=10\times f_{\text{hebb}}}$), as in \autoref{sec:benchmarking}.

\begin{figure}[htb]
    \centering
    \includegraphics[width=0.90\linewidth]{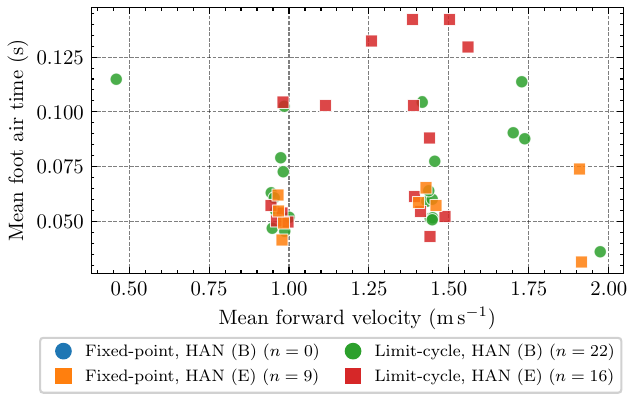}
    \caption{Mean foot air time versus forward velocity for HANs in conditions (B) and (E) across $60$ Unitree Go1 training runs, classified as fixed-point or limit-cycle attractors.}
    \label{fig:emerging_attractors}
\end{figure}

Consistent with the Gymnasium results (\autoref{sec:oscillations}), condition (B) produces oscillatory weight trajectories coupled to the gait cycle (\autoref{fig:go1_pca}, left), while condition (E) converges to stable fixed-point attractors after an initial adaptation phase (\autoref{fig:go1_pca}, right). Across $60$ training runs, condition (B) produced sustained locomotion in $22$ cases ($\SI{36.7}{\percent}$), all exhibiting limit-cycle dynamics. Condition (E) yielded stable locomotion in $25$ cases ($\SI{41.7}{\percent}$), of which $9$ ($\SI{36.0}{\percent}$) converged to fixed-point attractors and $16$ ($\SI{64.0}{\percent}$) remained limit-cycle; the remaining $13$ runs ($\SI{21.7}{\percent}$) did not produce stable locomotion. The parsimonious reward design — rewarding only target speed and yaw — allows diverse gaits to emerge, summarized via mean forward velocity and foot air time in \autoref{fig:emerging_attractors}, with attractors classified using the convergence criterion from \autoref{sec:oscillations} ($\rho=0.4$).

\subsection{Morphological adaptation}
\label{sec:ant_adaptation}
Having established that conditions (B) and (E) give rise to distinct attractor dynamics, we now ask whether these differences affect adaptive capabilities under morphological damage. Following Najarro et al.~\cite{najarro2020meta}, we evaluate adaptation capabilities on Ant-v5~\cite{towers2024gymnasium} by significantly shortening the distal shin segment of one of the four legs. We compare PPO against HANs condition (B) (HAN$^{M=1}_{f_{\text{NN}}=f_{\text{hebb}}}$) and condition (E) (HAN$^{M=20}_{f_{\text{NN}}=10\times f_{\text{hebb}}}$) across three training scenarios (no leg mutilation; front-right (FR) mutilated; front-left (FL) mutilated) and two unseen test scenarios (rear-left (RL) mutilated; rear-right (RR) mutilated) with an episode length of $500$.

\begin{figure}[!ht]
    \centering
    \includegraphics[width=0.97\linewidth]{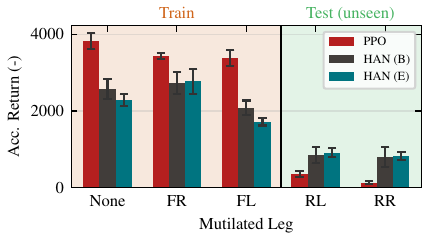}
    \caption{Mean accumulated returns with \SI{95}{\percent} confidence intervals for each scenario. The left side shows training scenarios, while the right side shows unseen test scenarios. PPO significantly outperforms HANs within the training distribution, while HANs achieve higher returns on unseen test scenarios.}
    \label{fig:ant_mutilated}
\end{figure}

PPO is trained for 1B time steps with default hyperparameters, 
and both HAN conditions for 6B time steps with $15$ random 
initializations using the same OpenAI-ES configuration as in 
\autoref{sec:quad}. Mutilations are presented to PPO by dividing 
the $1536$ training environments evenly, while HANs perform $4$ 
repeated fitness evaluations per mutilation during meta-training. 
We evaluate the final policy or Hebbian rules of each seed $30$ 
times per scenario (\autoref{fig:ant_mutilated}). 
PPO significantly outperforms HANs within the training distribution. 
Despite dual-timescale plasticity and averaging, all HAN runs 
converge to limit-cycle attractors in this setting, with no 
significant differences between conditions (B) and (E). 
Consistent with prior work~\cite{najarro2020meta, leung2025bio}, the 
oscillatory weight dynamics may facilitate adaptation, 
as HANs maintain more consistent performance across seen 
and unseen morphologies.
\section{Discussion}
\label{sec:discussion}\label{sec:conclusion}
In this work, we elucidate the effects of several Hebbian learning conditions on the dynamics of the weights.
While plain HNN can solve simple benchmarks, we find that introducing MN yields HANs that achieve significantly higher average performance on all investigated benchmarks, confirming previous studies~\cite{ferigo2022evolving, leung2025bio} on different locomotion problems. It might be expected that a lower HAN update frequency would reduce responsiveness and therefore performance. Contrary to this intuition, decoupling HAN updates results in improved performance over the given benchmarks. Furthermore, averaging synaptic activations can achieve higher average performance (see HalfCheetah and Hopper). On common locomotion benchmarks, decoupled HANs allow identification of operating conditions with emerging fixed-point attractors that show comparable or higher performance than co-dynamic instances and static NNs.
The timing of Hebbian updates strongly influences the evolution of Hebbian coefficients and gives rise to distinct plastic weight and attractor dynamics. A commonly used HAN configuration involving MN \cite{najarro2020meta, leung2025bio, schmidgall2020} introduces a limit cycle attractor with oscillating plastic weights in tandem with the periodic gait. 
This co-dynamic system drives the effective gait, as interrupting the Hebbian update during operational lifetime causes the agent to stumble (\autoref{fig:weight_freeze}, middle). These findings suggest that, in our setting, evolution tunes a co-dynamic system rather than creating a meta-learning mechanism.
More importantly, the introduction of different timescales and moving average with reduced Hebbian update frequency $f_{\text{hebb}}$ decouples the co-dynamics and results in the emergence of fixed-point attractors, which do not suffer from this weakness. In our perturbation test, neither freezing the weights nor applying a force perturbation disrupted the gait, suggesting robustness of the fixed-point attractor in this scenario. In the latter, the weights showed adaptation to regain stable locomotion before converging back to the same attractor (\autoref{fig:perturbation}). Under permanent morphological damage, both HAN conditions converge to limit-cycle attractors — even under the strong configuration of condition (E) — suggesting that evolution 
preferentially discovers oscillatory dynamics when adaptation across qualitatively different locomotion modes is required.
\section{Conclusion}
\label{sec:futurework}
This work provides a principled understanding of Hebbian learning and elucidates the role of dual-timescales and activation averaging. Specifically, weight analysis reveals the onset of different attractor dynamics, allowing us to design different adaptive properties in HANs. 
While the results highlight the potential of HANs for adaptive control, several limitations remain. ES, though effective for discovering plasticity rules, are sample-inefficient and have limited usability for complex tasks; more sophisticated optimization methods could expand applicability. We use feedforward networks, yet introducing recurrence may yield richer dynamics through interactions between Hebbian updates and recurrent activity. Furthermore, this paper centers on periodic locomotion; generalization of HANs to more complex tasks—such as navigation, manipulation, or complex terrains—remains an open question.
Future work will explore multi-attractor HANs, where autonomous transitions between attractors enable adaptation to sudden environmental changes. We hypothesize that such switching could be achieved by incorporating reward signals into the local Hebbian update, e.g. inspired by neuromodulation~\cite{soltoggio2007evolving, kudithipudi2022biological, soltoggio2018born}, helping bridge the sim-to-real gap. Finally, real-world experiments will be essential to assess HANs under noise, mechanical variation, and sensor drift.
\section*{Acknowledgements}
\label{sec:acknowledgements}
This work was supported by the Swiss National Science Foundation (SNSF) and the Japan Society for the Promotion of Science (JSPS) under project number \texttt{IZLJZ2\_214053}. Generative AI tools (OpenAI ChatGPT, Anthropic Claude) assisted with debugging, visualization scripts, and editorial improvements.




\bibliographystyle{IEEEtran}
\bibliography{references}


\newpage
\onecolumn
\raggedbottom
\appendices

\section{Method Details}
\label{sec:appendix_method}

\subsection{Hebbian update with dual-timescale plasticity and activation averaging}
\label{sec:code_hebb_update}
\begin{center}
\begin{minipage}{0.65\linewidth}
\begin{algorithm}[H]
\caption{Hebbian update with Activation Buffer}
\label{al:optimization_flow}
\begin{algorithmic}[1]
\State Initialize policy with parameters $\theta_{W}$, activation buffer $\mathcal{D}_{syn}$
\For{episode = 1 to $N_{episodes}$}
    \For{$t = 1$ to $T$}
        \State Observe state $s_t$; compute action $a_t = \pi_{\theta_W}(s_t)$
        \State Apply $a_t$ in environment; observe $r_t$, $s_{t+1}$
        \State Store $(x_j^{(k-1)}(t), x_i^{(k)}(t))$ in $\mathcal{D}_{syn}$
        \If{$t \bmod t_{\text{update}} = 0$}
            \State Compute smoothed $\hat{x}_j^{(k-1)}(t), \hat{x}_i^{(k)}(t)$ over $\mathcal{D}_{syn}$
            \State Compute $\Delta w_{ij}^{(k)}(t) = \eta_{ij}^{(k)} \cdot h_{\theta_{ij}^{(k)}}\!\left(\hat{x}_j^{(k-1)}(t), \hat{x}_i^{(k)}(t)\right)$
            \State Update $W \leftarrow W + \Delta W$
        \EndIf
    \EndFor
\EndFor
\end{algorithmic}
\end{algorithm}
\end{minipage}
\end{center}

\subsection{Matrix formulation of Hebbian updates}
\label{appendix:matrix_form}
For computational efficiency, the Hebbian updates are implemented in matrix form using elementwise operations. The update for all weights in layer $k$ can be written as:
\begin{equation}
\Delta W^{(k)} = \eta^{(k)} \odot \left(
A^{(k)} \odot \left(\hat{\boldsymbol{x}}^{(k-1)} \otimes \hat{\boldsymbol{x}}^{(k)}\right)
+ B^{(k)} \otimes \hat{\boldsymbol{x}}^{(k-1)}
+ C^{(k)} \otimes \hat{\boldsymbol{x}}^{(k)}
+ D^{(k)}
\right) \nonumber
\end{equation}
where $\odot$ denotes the Hadamard (elementwise) product, $\otimes$ denotes the outer tensor product, and $A^{(k)}, B^{(k)}, C^{(k)}, D^{(k)} \in \mathbb{R}^{n_k \times n_{k-1}}$ are matrices of Hebbian coefficients. The learning rates $\eta^{(k)}$ are stored per connection as a matrix of the same shape. The smoothed activations $\hat{\boldsymbol{x}}^{(k)}$ and $\hat{\boldsymbol{x}}^{(k-1)}$ are broadcast to align with the weight dimensions as needed.

\subsection{Evolutionary algorithm}
\label{appendix:evolution}
We utilize the \texttt{SimpleES} implementation of \texttt{evosax} (Version \texttt{0.1.6}).
\begin{center}
\begin{minipage}{0.7\linewidth}
\begin{algorithm}[H]
\caption{Evolution Strategy with step size adaptation (AdaptiveES)}
\begin{algorithmic}[1]
\State \textbf{Input:} population size $N$, number of parameters $d$, mean step size $c_\mu$, mutation step size $c_\sigma$, initial mutation rate $\sigma_{init}$
\Procedure{Initialize}{} 
    \State mean $\mu \sim \mathcal{U}(-0.1, 0.1)^d$, weights $w_i = \frac{1}{N_{\text{parents}}}$
\EndProcedure
\Procedure{Ask}{$\mu, \sigma$}
    \State Sample $z_i \sim \mathcal{N}(0, I)$ for $i = 1, \dots, N$
    \State $x_i \gets \mu + \sigma \cdot z_i$
    \State \Return population $\{x_i\}$
\EndProcedure
\Procedure{Tell}{$\{x_i\}, \{f_i\}, \mu, \sigma$}
    \State Evaluate fitness $f_i$ for each $x_i$
    \State Select top $N_{\text{parents}}$ individuals by fitness
    \State Compute weighted update: $y_w = \sum w_i (x_i - \mu)$
    \State Update mean: $\mu \gets \mu + c_\mu \cdot y_w$
    \State Estimate step size: $\hat{\sigma} \gets \sqrt{\sum w_i (x_i - \mu)^2}$
    \State Update $\sigma \gets (1 - c_\sigma)\sigma + c_\sigma \hat{\sigma}$
    \State \Return updated $\mu, \sigma$
\EndProcedure
\end{algorithmic}
\end{algorithm}
\end{minipage}
\end{center}
Unlike vanilla ES, this algorithm uses an adaptive mutation rate based on the selected parent generation. In preliminary tests, we find that this algorithm is robust to hyperparameter choices and requires few hyperparameters.

\section{Experimental Details}
\label{sec:appendix_experimental}

\subsection{Hyperparameters Gymnasium benchmarks}
\label{sec:gymnasium_hyperparams}
\begin{table}[H]
\centering
\caption{Hyperparameters of controller evolution.}
\begin{tabular}{rl}
\hline
\multicolumn{1}{c}{\textbf{Hyperparameter}} & \multicolumn{1}{c}{\textbf{Value}} \\ \hline
\multicolumn{2}{c}{\textbf{Training settings}} \\ \hline
Number of generations      & $500$ (Swimmer) / $1000$ (rest) \\
Repeats                    & $4$ \\
Action clipping            & $-1, 1$ \\
Observation normalization  & \begin{tabular}[c]{@{}l@{}}Running standard normalization \\ (Chan's online algorithm)\end{tabular} \\ \hline
\multicolumn{2}{c}{\textbf{Evolutionary Algorithm}} \\ \hline
Population size            & $128$ \\
Initial mutation rate $\sigma_{init}$ & $0.5$ \\
Step size mean $c_\mu$     & $1.0$ \\
Step size mutation $c_\sigma$ & $0.1$ \\
Selection ratio            & \SI{10}{\percent} \\ \hline
\multicolumn{2}{c}{\textbf{Static Neural Network Architecture}} \\ \hline
Hidden activation function & $\tanh$ \\
Output activation function & $\tanh$ \\
Weight initialization      & $\mathcal{U}(-0.1, 0.1)$ \\ \hline
\multicolumn{2}{c}{\textbf{Hebbian Neural Network Architecture}} \\ \hline
Hidden neurons             & $16$ \\
Hidden activation function & $\tanh$ \\
Output activation function & $\tanh$ \\
(Plastic) weight initialization & $\mathcal{U}(-0.1, 0.1)$ \\
Hebbian coefficient initialization & $\mathcal{U}(-1.0, 1.0)$ \\
\hline
\end{tabular}
\end{table}

\begin{table}[H]
\centering
\caption{Number of trainable parameters for different neural controllers across Gymnasium benchmarks.}
\label{tab:trainable_params}
\begin{tabular}{@{}lrrrrrr@{}}
\toprule
Controller & Architecture & Bias & Swimmer-v5 & Hopper-v5 & HalfCheetah-v5 & Walker2d-v5 \\ \midrule
HNN / HAN & \texttt{MLP} (16) & \xmark & 800 & 1120 & 1840 & 1840 \\
PPO tiny & \texttt{MLP} (16) & \cmark & 178 & 243 & 390 & 390 \\
PPO default & \texttt{MLP} (64, 64) & \cmark & 4866 & 5123 & 5702 & 5702 \\ 
ES MLP & \texttt{MLP} (80) & \xmark & 800 & 1120 & 1840 & 1840 \\
ES MLP & \texttt{MLP} (variable) & \cmark & 805 & 1113 & 1830 & 1830 \\
ES GRU & \texttt{GRU} (variable) & \cmark & 782 & 1137 & 1893 & 1893 \\
\bottomrule
\end{tabular}
\end{table}

\begin{table}[H]
\centering
\caption{Default hyperparameters for PPO.}
\label{tab:ppo-hyperparams}
\begin{tabular}{lc}
\toprule
\textbf{Hyperparameter} & \textbf{Value} \\
\midrule
Learning rate & 3e-4 \\
Number of steps per rollout ($n_{\text{steps}}$) & 2048 \\
Minibatch size & 64 \\
Number of epochs & 10 \\
Discount factor ($\gamma$) & 0.99 \\
GAE lambda & 0.95 \\
Clip range (actor) & 0.2 \\
Clip range (critic) & - \\
Normalize advantage & True \\
Entropy coefficient ($c_{\text{ent}}$) & 0.0 \\
Value function coefficient ($c_{\text{vf}}$) & 0.5 \\
Max gradient norm & 0.5 \\
Target KL divergence & None \\
Critic architecture & \texttt{MLP} (2 layers, 64 units each) \\
\bottomrule
\end{tabular}
\end{table}

\subsection{Unitree Go1 locomotion}
\label{sec:setting_unitree_go1}
We train HAN controllers on a velocity-tracking task using a simulated Unitree Go1 robot. The controller is a single-hidden-layer \texttt{MLP} with 16 neurons and $\tanh$ activations. Hebbian plasticity follows an evolved $\eta$-ABCD rule. Optimization is performed via OpenAI-ES~\cite{salimans2017evolution} with a population size of $512$ and exponential decay schedules for learning rate and mutation strength. Rollouts are not terminated early; all episode steps contribute to the fitness.

\begin{table}[ht!]
\centering
\caption{Evolutionary algorithm parameters for Unitree Go1 locomotion.}
\begin{tabular}{ll}
\hline
\textbf{Parameter} & \textbf{Value} \\
\hline
Population Size & $512$ \\
\midrule
Optimizer & Adam \\
\quad Momentum coefficient ($\beta_1$) & $0.9$ \\
\quad Second moment coefficient ($\beta_2$) & $0.999$ \\
\quad Epsilon & $1 \times 10^{-8}$ \\
\midrule
Learning Rate Schedule & Exponential Decay \\
\quad Initial Learning Rate & $0.1$ \\
\quad Decay Rate & $0.999$ \\
\midrule
Mutation Rate Schedule & Exponential Decay \\
\quad Initial Mutation Rate & $0.2$ \\
\quad Decay Rate & $0.995$ \\
\midrule
Fitness Shaping & Center Ranking \\
\hline
\end{tabular}
\label{tab:config_go1}
\end{table}

\begin{figure}[H]
    \centering
    \includegraphics[width=0.6\linewidth]{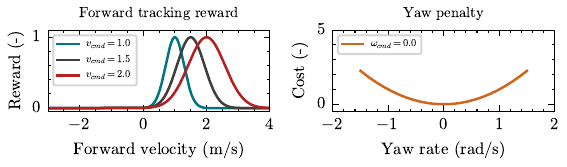}
    \caption{Unitree Go1 reward function components.}
    \label{fig:unitree_reward}
\end{figure}

The locomotion task is adapted from the Joystick task of MuJoCo Playground. The reward function consists of a Gaussian tracking term for a target forward velocity and a Gaussian tracking term for maintaining zero yaw (\autoref{fig:unitree_reward}). We use $4$ repeats per evaluation and simulate each episode for $1000$ steps at a controller frequency of \SI{50}{\hertz}. A full breakdown of all physics parameters is available in the published code on the project webpage.

\section{Learning Curves}
\label{sec:appendix_learning_curves}

\begin{figure}[H]
    \centering    
    \includegraphics[width=0.62\linewidth]{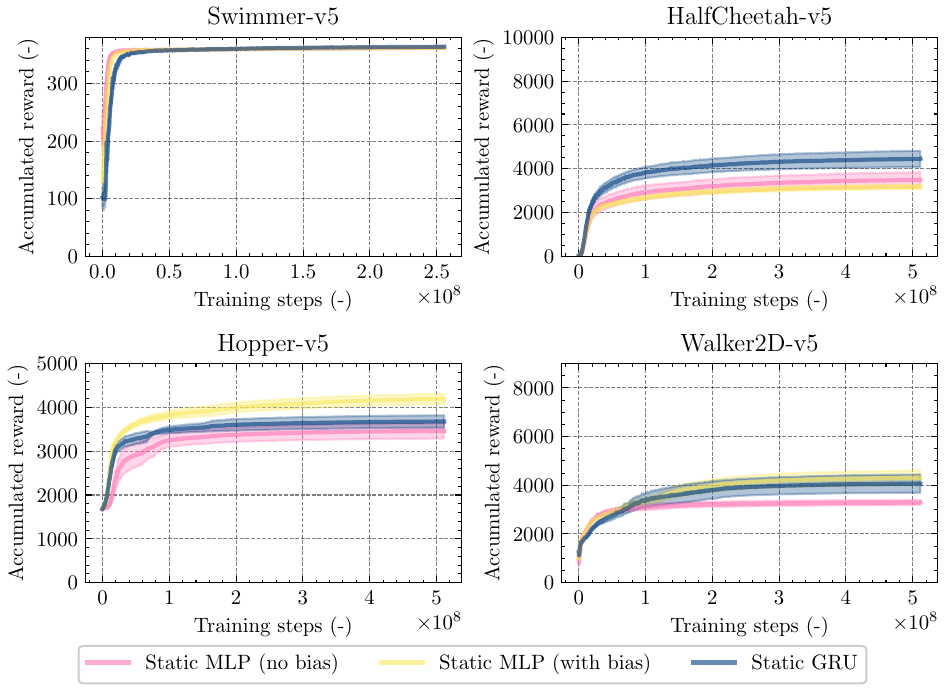}
    \caption{Mean and \SI{95}{\percent} confidence intervals of 4 independent rollouts of $n=16$ randomly initialized training runs of static neural networks with ES.}
    \label{fig:benchmarks_static}
\end{figure} 
\begin{figure}[H]
    \centering    
    \includegraphics[width=0.62\linewidth]{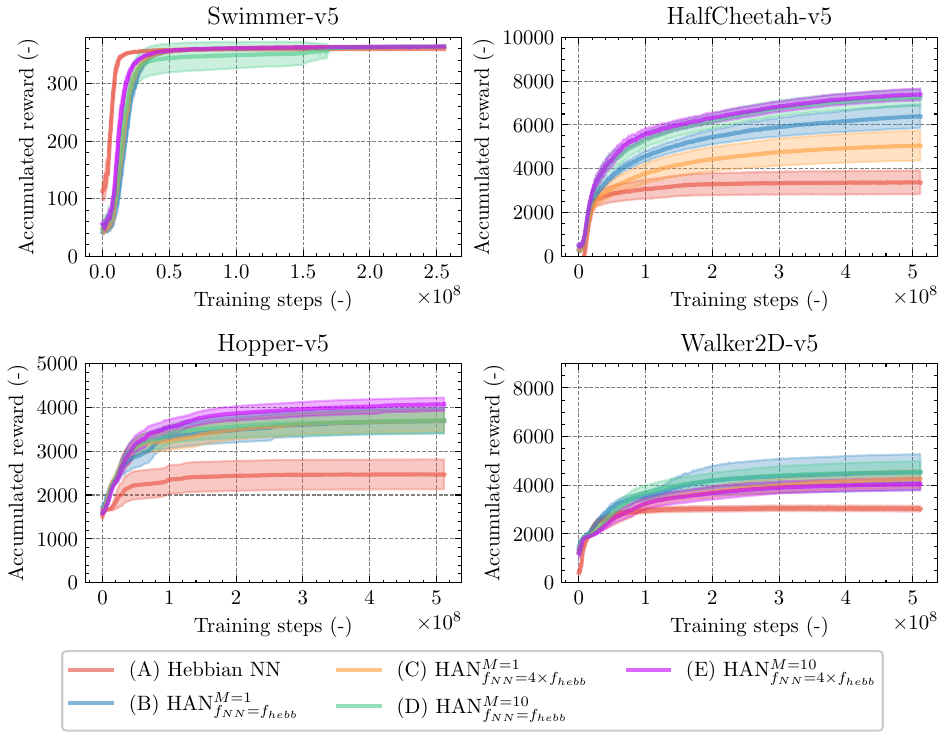}
    \caption{Mean and \SI{95}{\percent} confidence intervals of 4 independent rollouts of $n=16$ randomly initialized meta-training runs of Hebbian update rules in different conditions with ES.}
    \label{fig:benchmarks_hebbian}
\end{figure} 
\begin{figure}[H]
    \centering    
    \includegraphics[width=0.62\linewidth]{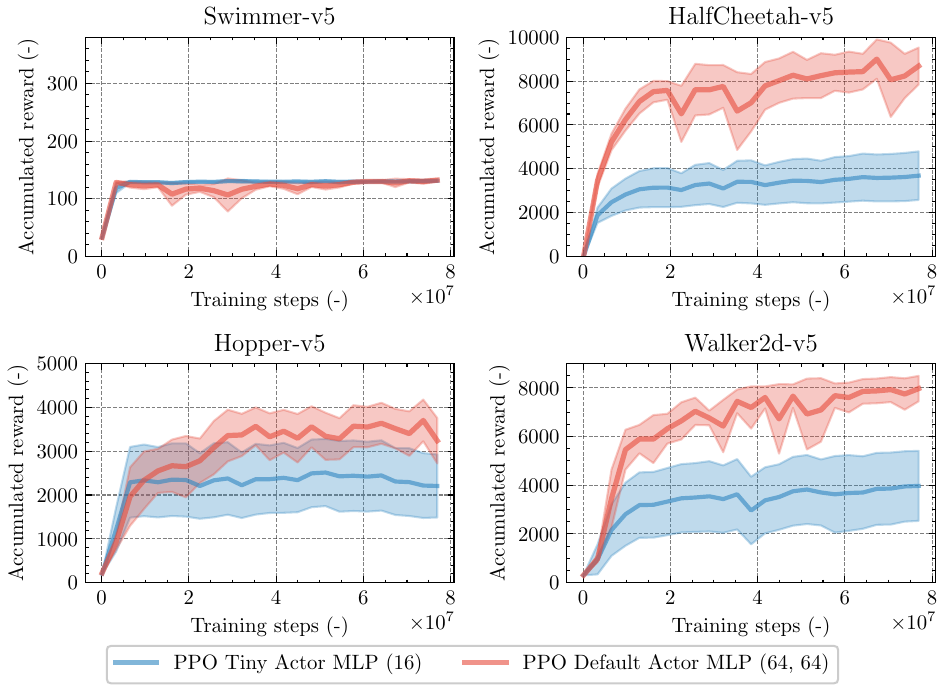}
    \caption{Mean and \SI{95}{\percent} confidence intervals of $n=10$ randomly initialized training runs with PPO.}
    \label{fig:benchmarks_ppo}
\end{figure} 

\newpage
\section{Additional Weight Dynamics Analyses}
\label{sec:appendix_weight_dynamics}

\subsection{Weight dynamics without max normalization}
\label{sec:weight_dynamics_no_mn}
\begin{figure}[H]
    \centering
    \includegraphics[width=0.7\linewidth]{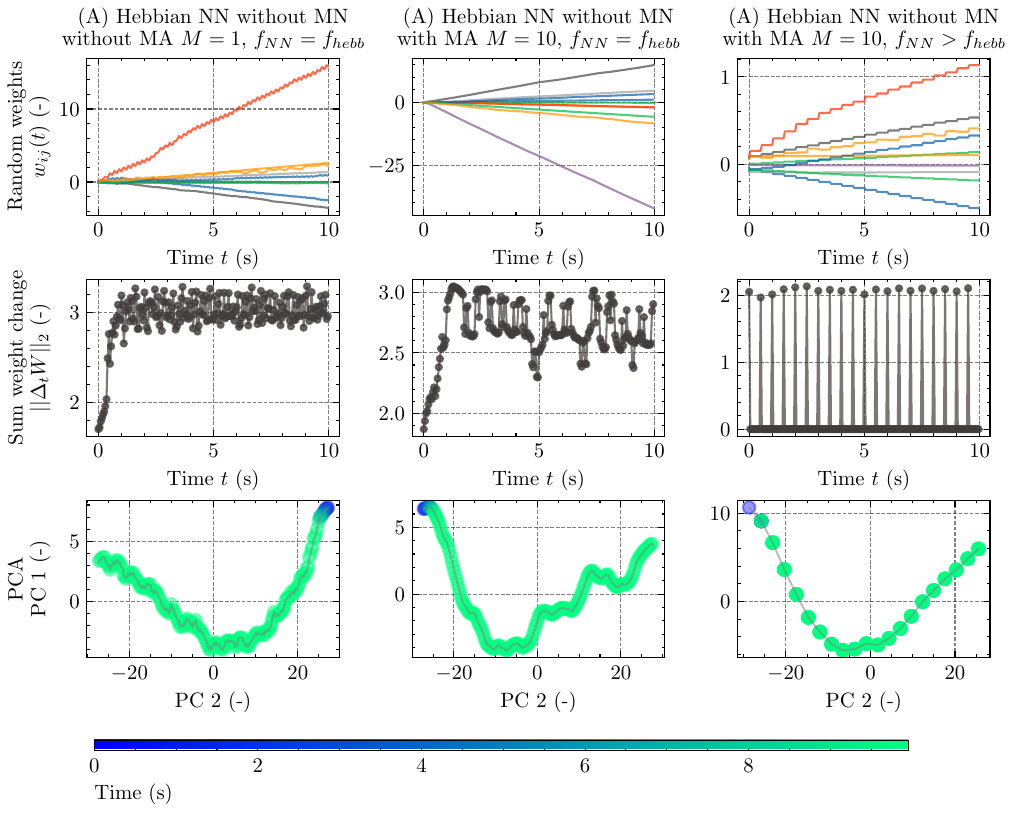}
    \caption{Weight dynamics of HalfCheetah-v5 without max normalization.}
    \label{fig:weight_dynamics_without_ws}
\end{figure}

\autoref{fig:weight_dynamics_without_ws} shows the weight dynamics when no max normalization is applied. Unlike with max normalization, we observe unbounded growth and no distinct attractor behavior in all three conditions.

\subsection{Random weight initializations}
\label{sec:random_init}
\begin{figure}[hbt]
    \centering
    \includegraphics[width=1\linewidth]{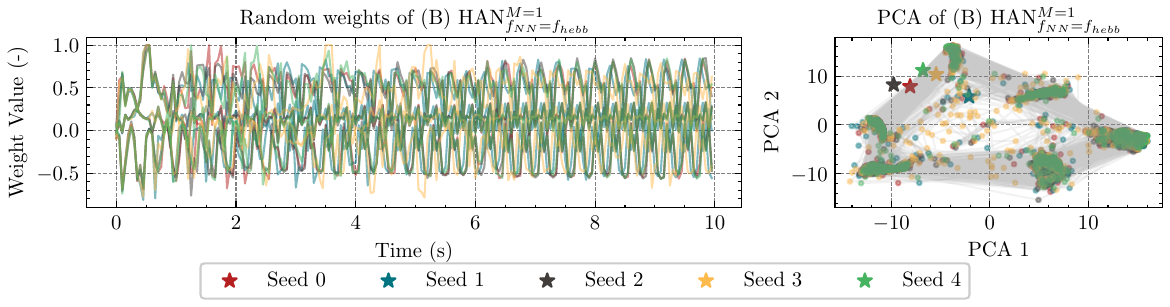}
    \caption{Weight dynamics across five random weight initializations of HAN$^{M=1}_{f_\text{NN}=f_\text{hebb}}$ (limit-cycle attractor). \textbf{Left}: Time evolution of three selected plastic weights. \textbf{Right}: Projection onto the first two global PCA components reveals consistent oscillatory patterns across seeds.}
    \label{fig:random_weights_lca}
\end{figure}

\begin{figure}[H]
    \centering
    \includegraphics[width=1\linewidth]{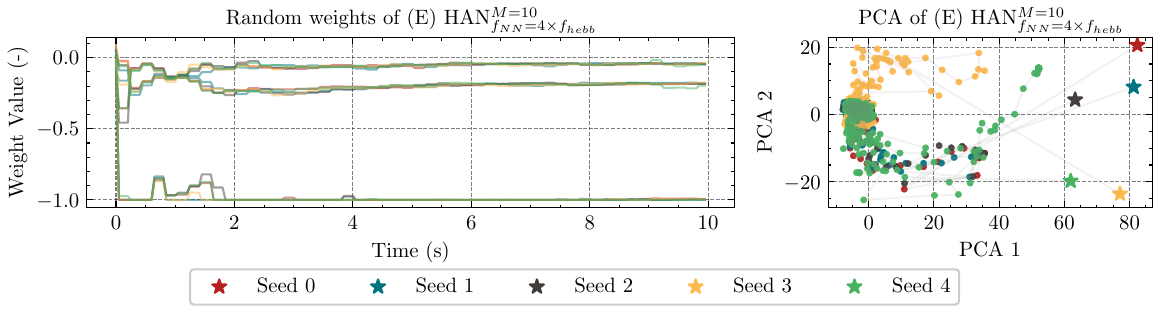}
    \caption{Weight dynamics across five random weight initializations of HAN$^{M=10}_{f_\text{NN}=4\times f_\text{hebb}}$ (fixed-point attractor). \textbf{Left}: Time evolution of three selected plastic weights. \textbf{Right}: PCA projection shows rapid convergence to a compact region in weight space regardless of initial weights.}
    \label{fig:random_weights_fpa}
\end{figure}

To assess the sensitivity of weight dynamics to the initial network configuration, we initialize HANs with different random seeds and compare the resulting weight trajectories. We use the same Hebbian coefficients analyzed in \autoref{sec:oscillations} and track three randomly chosen plastic weights under five distinct network initializations. The environment seed is held constant to isolate the effect of weight initialization. We compute a shared PCA basis from concatenated weight trajectories across all seeds and project each trajectory onto this global embedding. \autoref{fig:random_weights_lca} shows the weight dynamics under a limit-cycle attractor, while \autoref{fig:random_weights_fpa} shows the same analysis for a fixed-point attractor.

\section{Ablation Studies}
\label{sec:appendix_ablations}

\subsection{Oja's rule and weight update clipping}
\label{sec:appendix_oja_clipping}
\begin{figure}[htb]
    \centering
    \includegraphics[width=1\linewidth]{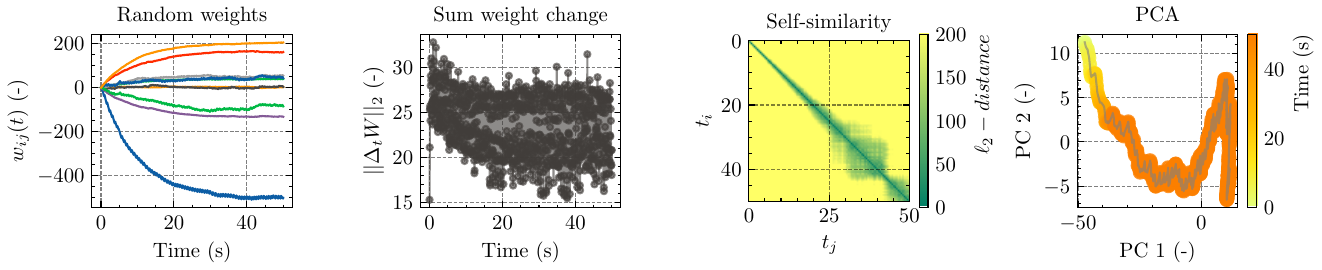}
    \caption{Weight dynamics of Oja-regularized HNNs during deployment.
        \textbf{(1)}: Selected plastic weights remain bounded and non-periodic.
        \textbf{(2)}: Summed weight change decreases initially and stabilizes with high oscillations.
        \textbf{(3)}: $\ell_2$-distance between time instances remains high.
        \textbf{(4)}: PCA projection shows non-oscillatory, continuous drift.}
    \label{fig:oja_dynamics}
\end{figure}

We explore two alternatives to max normalization for preventing unbounded plastic weight growth:
\begin{itemize}
    \item \textbf{Weight update clipping} — elementwise bounding of Hebbian updates.
    \item \textbf{Oja's rule} — a decay term proportional to both the weight and postsynaptic activation.
\end{itemize}
The clipping operation constrains the magnitude of each Hebbian update:
\begin{equation}
\label{eq:clipping}
\Delta w_{ij}^{(k)} = \text{clip} \left( \Delta w_{ij}^{(k)},\ -\varepsilon,\ \varepsilon \right),
\end{equation}
with thresholds $\varepsilon \in \{1, 2, 5, 10, 100, 1000, 10000\}$. Tight clipping limits the learning signal, whereas loose thresholds permit divergence.
\begin{figure}[H]
    \centering
    \includegraphics[width=0.95\linewidth]{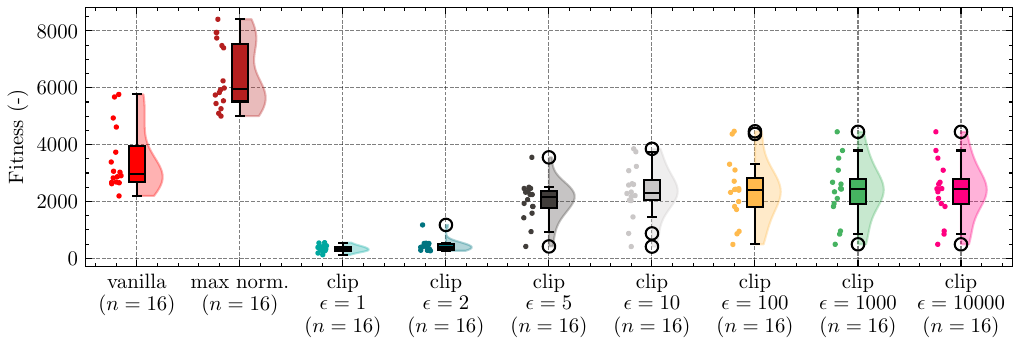}
    \caption{Fitness of HNNs on HalfCheetah-v5 with weight clipping at various thresholds $\varepsilon$. Moderate clipping ($\varepsilon=5$) preserves learning, but max normalization outperforms all clipped variants.}
    \label{fig:raindrop_clipping}
\end{figure}
We also evaluate Oja's rule~\cite{oja1982simplified}, which introduces a decay term proportional to the postsynaptic activity and the current weight:
\begin{equation}
\label{eq:oja}
\Delta w_{ij}^{(k)} \leftarrow \Delta w_{ij}^{(k)} - \eta_{ij}^{(k)} \cdot \left( x_i^{(k)} \right)^2 \cdot w_{ij}^{(k)}.
\end{equation}
We test learning rates $\eta \in \{0.005, 0.01, 0.05, 0.1\}$.
\begin{figure}[H]
    \centering
    \includegraphics[width=0.75\linewidth]{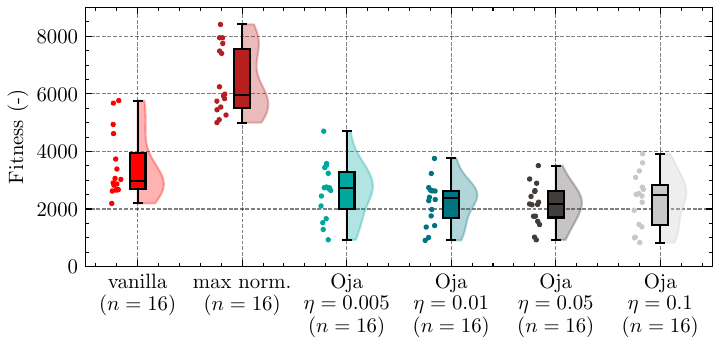}
    \caption{Fitness across 16 evolutionary runs for different regularization methods. Oja's rule does not yield competitive performance compared to max normalization.}
    \label{fig:raindrop_oja}
\end{figure}
\autoref{fig:oja_dynamics} shows the weight dynamics of Oja-regularized HNNs. While the decay mechanism constrains weights, the total plasticity remains high due to large individual weight values, and no stable attractor behavior emerges.
Both clipping and Oja's rule constrain weight growth, but neither matches the performance of max normalization.

\subsection{Constant vs.\ evolved Hebbian learning rate}
\label{sec:appendix_learning_rate}
We evaluate whether co-evolving the Hebbian learning rate offers benefits over fixed values on HalfCheetah-v5, using the ABCD rule with max normalization and 16 hidden neurons. We test constant learning rates $\eta \in \{0.005, 0.01, 0.05, 0.1, 0.5, 1.0\}$ alongside a co-evolved variant, which includes learning rates in the evolutionary search space (\SI{20}{\percent} increase in parameter count).
\begin{figure}[H]
    \centering
    \includegraphics[width=0.85\linewidth]{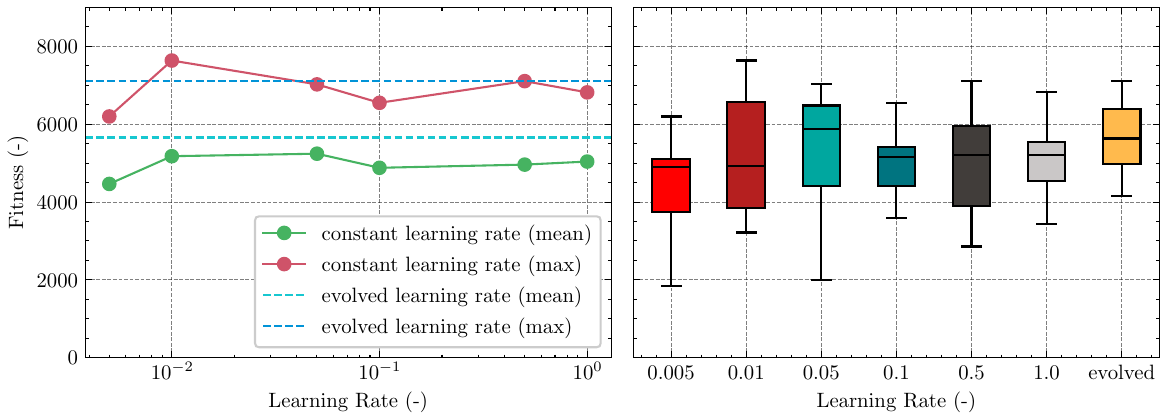}
    \caption{Best-performing individual from each of 16 evolutionary runs on HalfCheetah-v5 for constant and co-evolved Hebbian learning rates. No clear advantage is observed for evolving the learning rate.}
    \label{fig:const_learning_rate}
\end{figure}
Co-evolved learning rates show a slight decrease in variance, but no significant advantage over manually selected values.

\subsection{Neural network size}
\label{sec:network_size}
We evaluate the effect of network architecture size on HalfCheetah-v5. \autoref{tab:sizes} lists the evaluated architectures.
\begin{table}[H]
    \centering
    \caption{Neural network architectures by size and parameter count.}
    \begin{tabular}{c c c c}
        \toprule
        Size & \begin{tabular}[c]{@{}c@{}}Hidden layer\\ size\end{tabular} & 
               \begin{tabular}[c]{@{}c@{}}Optimization\\ parameters\end{tabular} & 
               \begin{tabular}[c]{@{}c@{}}Network\\ parameters\end{tabular} \\
        \midrule
        \texttt{XXXS}     & 2              & 184    & 46   \\
        \texttt{XXS}      & 4              & 368    & 92   \\
        \texttt{XS}       & 8              & 736    & 184  \\
        \texttt{S}        & 16             & 1472   & 368  \\
        \texttt{M}        & 32             & 2944   & 736  \\
        \texttt{L}        & 32, 32         & 7040   & 1760 \\
        \texttt{XL}       & 64, 32         & 13312  & 3328 \\
        \texttt{XL deep}  & 32, 32, 32, 32 & 15232  & 3808 \\
        \texttt{XXL}      & 128, 32        & 25856  & 6464 \\
        \bottomrule
    \end{tabular}
    \label{tab:sizes}
\end{table}
\begin{figure}[h]
    \centering
    \includegraphics[width=0.95\linewidth]{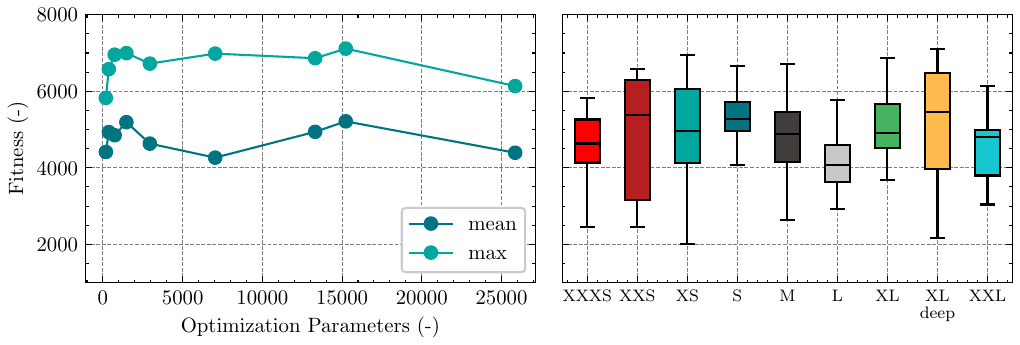}
    \caption{Performance of 16 evolutionary runs per network size on HalfCheetah-v5.
        \textbf{Left}: Mean and maximum performance vs.\ optimization parameter count.
        \textbf{Right}: Fitness distribution across seeds for each architecture.}
    \label{fig:architecture_size_boxplot}
\end{figure}
Network size has a limited effect on overall performance. Very small (\texttt{XXXS}) and very large (\texttt{XXL}) architectures achieve slightly lower fitness on average, but intermediate sizes perform comparably.

\newpage
\section{Switching Attractors}
\label{sec:switching-attractors}

\begin{figure}[H]
    \centering
    \begin{subfigure}{0.49\textwidth}
        \centering
        \includegraphics[width=0.9\textwidth]{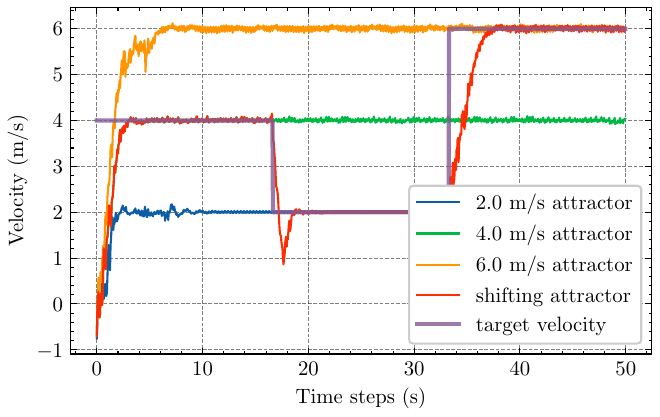}
        \caption{Forward velocity during attractor switching.}
        \label{fig:switching_attractors_ts}
    \end{subfigure}
    \hspace{0.0cm}
    \begin{subfigure}{0.48\textwidth}
        \centering
        \includegraphics[width=0.9\textwidth]{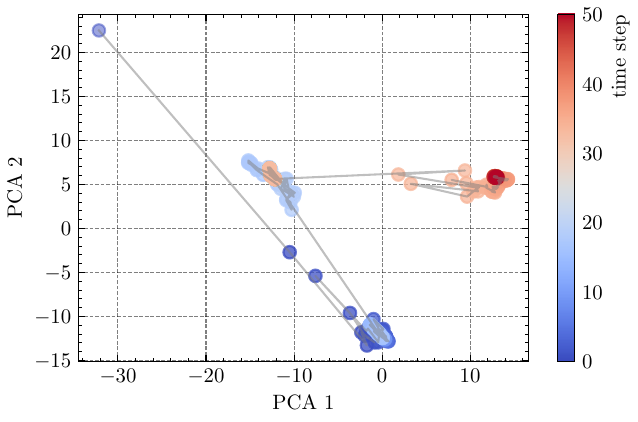}
        \caption{PCA of weight trajectories during switching.}
        \label{fig:switching_attractors_pca}
    \end{subfigure}
    \caption{Attractor switching on HalfCheetah-v5 during deployment. Hebbian coefficients are swapped at \SI{16.6}{\second} and \SI{33.3}{\second}, causing the network to transition between different fixed-point attractors.}
    \label{fig:switching_attractors}
\end{figure}

As a preliminary investigation, we test whether HANs can switch between different fixed-point attractors during deployment by loading different Hebbian coefficients on-the-fly. We evolve separate HAN$^{M=10}_{f_{\text{NN}}=4 \times f_{\text{hebb}}}$ controllers for a modified HalfCheetah-v5 task with a velocity tracking reward $r_t = (v_{\text{target}} - v_x)$ at three target velocities: \SIlist{2;4;6}{\meter\per\second}. Plastic network weights are initialized from $\mathcal{U}(-1.0, 1.0)$ to cover the achievable parameter space under max normalization. Each controller is evolved over $1000$ generations using the same training setting as in \autoref{sec:indepth}.

During deployment, we initialize with the \SI{4}{\meter\per\second} coefficients, switch to the \SI{2}{\meter\per\second} coefficients at \SI{16.6}{\second}, and to the \SI{6}{\meter\per\second} coefficients at \SI{33.3}{\second}. \autoref{fig:switching_attractors} shows the resulting velocity tracking and PCA projections of the weight trajectories.

\end{document}